\documentclass[utf8]{FrontiersinHarvard}

\usepackage{tikz}
\usetikzlibrary{calc, positioning, spy}

\usepackage{standalone}

\usepackage{multirow}

\usepackage[T1]{fontenc} 
\usepackage{lmodern}

\usepackage[algo2e]{algorithm2e}

\usepackage{amsmath}
\usepackage{booktabs}

\newif\iftrailingfigures
\trailingfigurestrue
\trailingfiguresfalse 

\def\edhucat{EDHuCAT}

\def\rawtitle{AMaze: An intuitive benchmark generator for fast prototyping of generalizable agents}

\graphicspath{{annexes}{annexes_paper}}

\usepackage{lineno}

\newenvironment{localtoimage}[3][]{%
 \node (I) [inner sep=0pt] {\includegraphics[#3]{#2}};%
 \begin{localto}[#1]{I}%
}{%
 \end{localto}%
}
\newenvironment{localto}[2][]{
 \path let \p1=(#2.center), \p2=(#2.north), \p3=(#2.east) in coordinate (#2X) at (\x3-\x1,0) coordinate (#2Y) at (0,\y2-\y1);
 \begin{scope}[x={(#2X)}, y={(#2Y)}, shift={(#2.center)},#1]
}{
 \end{scope}
}

\usepackage{hyperref}

\def\firstAuthorLast{Kevin Godin-Dubois {et~al.}}
\def\Authors{K. Godin-Dubois\,$^{1,*}$, K. Miras\,$^{1}$ and A. V. Kononova\,$^{2}$}

\def\Address{
 $^{1}$Vrije Universiteit Amsterdam, Amsterdam, Netherlands \\
 $^{2}$Leiden University, Leiden, Netherlands
}

\def\corrAuthor{Kevin Godin-Dubois}
\def\corrEmail{k.j.m.godin-dubois@vu.nl}

\begin{document}

\onecolumn

\begin{frontmatter}
\onecolumn
\firstpage{1}

\title[AMaze: a benchmark generator for sighted agents]{\rawtitle}

\author[\firstAuthorLast ]{\Authors}
\address{}
\correspondance{}

\extraAuth{}

\maketitle


\begin{abstract}
 Traditional approaches to training agents have generally involved a single, deterministic environment of minimal complexity to solve various tasks such as robot locomotion or computer vision.
However, agents trained in static environments lack generalization capabilities, limiting their potential in broader scenarios.
Thus, recent benchmarks frequently rely on multiple environments, for instance, by providing stochastic noise, simple permutations, or altogether different settings.
In practice, such collections result mainly from costly human-designed processes or the liberal use of random number generators.
In this work, we introduce AMaze, a novel benchmark generator in which embodied agents must navigate a maze by interpreting visual signs of arbitrary complexities and deceptiveness.
This generator promotes human interaction through the easy generation of feature-specific mazes and an intuitive understanding of the resulting agents' strategies.
As a proof-of-concept, we demonstrate the capabilities of the generator in a simple, fully discrete case with limited deceptiveness.
Agents were trained under three different regimes (one-shot, scaffolding, interactive), and the results showed that the latter two cases outperform direct training in terms of generalization capabilities.
Indeed, depending on the combination of generalization metric, training regime, and algorithm, the median gain ranged from 50\% to 100\% and maximal performance was achieved through interactive training, thereby demonstrating the benefits of a controllable human-in-the-loop benchmark generator.

\tiny
 \def\keyFont{\fontsize{8}{11}\helveticabold }
 \keyFont{\section{Keywords:} Benchmark, Human-in-the-loop, Generalization, Mazes, Reinforcement Learning}
\end{abstract}
\end{frontmatter}

\section{Introduction}

Based on the need to fairly compare algorithms \citep{Islam2018}, benchmarks have proliferated in the Reinforcement Learning (RL) community.
These cover a wide range of tasks, from the full collection of Atari 2600 games \citep{Bellemare2013} to 3D simulations in Mujoco \citep{Laskin2021}.
However, in recent years, the focus of research has changed from producing more complex environments to producing a \emph{range} of environments.
Although undeniable progress has been made with respect to the capabilities of trained agents, much remains to be done for their capacity to generalize \citep{Mnih2015}.
In practice, agents ``will not learn a general policy, but instead a policy that will only work for a particular version of a particular task with particular initial parameters''~\citep{Risi2020}.

Thus, a recurring theme in modern RL research is the training of agents in various situations to avoid overfitting.
Although some algorithms have built-in solutions to smooth out the learning process, e.g. TD3 \citep{Fujimoto2018} (where small perturbations are applied to the actions), providing such a diversity of experience primarily originates from the environments themselves.
To this end, numerous benchmarks now consist of a collection with varying degrees of homogeneity.
Some of them have a similar structure, as in the Sonic benchmark \citep{Nichol2018} where levels are small areas taken from three games in the franchise.
In other cases, environments share very little: In the Arcade Learning Environment (ALE), the single common factors are the dimensions of the observation space \citep{Bellemare2013}.
Intermediate test suites with distinct but complementary sets of ``skill-building'' tasks have also been designed, for example, with the Mujoco simulator \citep{Wawrzynski2009, Yu2019, Laskin2021} or Meta-World \cite{Yu2019}.
However, all of these examples share a common feature: the set of environments is predefined, generally the result of a costly human-tailored design procedure, e.g. \citep{Beattie2016}.

To solve this generalization problem, agents must face sufficiently diverse situations so that the underlying principles are learned instead of a specific trajectory.
Naturally, this requires generating environments that exhibit such diversity while still offering the same core challenges.
One common way to address this later point is to use procedural generation \citep{Beattie2016,Kempka2016,Harries2019,Juliani2019,Tomilin2022} or complementary techniques such as evolutionary algorithms \citep{Alaguna2018,Wang2019}.
For example, ProcGen \citep{Cobbe2020} encompasses 16 different types of environment and serves as a generalizable alternative to ALE.
Adapting more recent video game environments, either directly \citep{Synnaeve2016} or in a light format \citep{Tian2017}, can help further push adaptability by providing finer-grained perceptions and actions.
While such an approach can be used to create large training sets, the main difficulty becomes the design of a sufficiently tunable generator, i.e. one in which desirable features are easy to introduce.

Considering the challenges of generating a panel of demanding training environments, the contribution of this article is two-fold:
\begin{enumerate}
 \item We introduce AMaze\footnote{AMaze library is available on PyPI at \url{https://pypi.org/project/amaze-benchmarker/}. The code for the experiment described thereafter is hosted at \url{https://github.com/kgd-al/amaze_edhucat_2024}}, a generator for generic, computationally inexpensive environments of unbounded complexity that focus on generalization (via environmental diversity) and intelligibility (intuitive human understanding).
 \item We demonstrate how such a generator is helpful in leading to more generalized performance (robust behavior w.r.t. unseen tasks) and how it can benefit from human input (e.g., to dynamically adjusting difficulty).
\end{enumerate}

After highlighting, in \autoref{sec:litterature}, the niche this generator occupies in the current benchmark literature, we describe its main components in \autoref{sec:mazes}.
Three alternative methodologies for training generalized maze-navigating agents are then detailed in \autoref{sec:training} alongside two algorithms (A2C and PPO).
The resulting performance in handling unknown environments is then thoroughly tested in \autoref{sec:eval}, allowing us to draw conclusions about the relative benefits of the generator, the training processes, and the underlying algorithms.

\section{Related benchmarks}
\label{sec:litterature}

\def\tabbenchmarks{
\begin{table}
 \caption{
  Comparison between AMaze and related benchmark (suites).
  All time metrics correspond to the wall time for 1000 timesteps of the corresponding environment, averaged over 10 replicates.
  Qualification of the inputs, outputs, and control levels are taken from the related article or, when unavailable, directly from the sources.
  Overall, AMaze is competitive with small-scale benchmarks, but provides the experimenter with more control over the characteristics of the targeted environments.
  Complex environments (e.g. 3D) have much higher computational costs making AMaze an efficient and scalable prototyping platform.
 }
 \label{tab:benchmarks}
 \includegraphics[width=\textwidth]{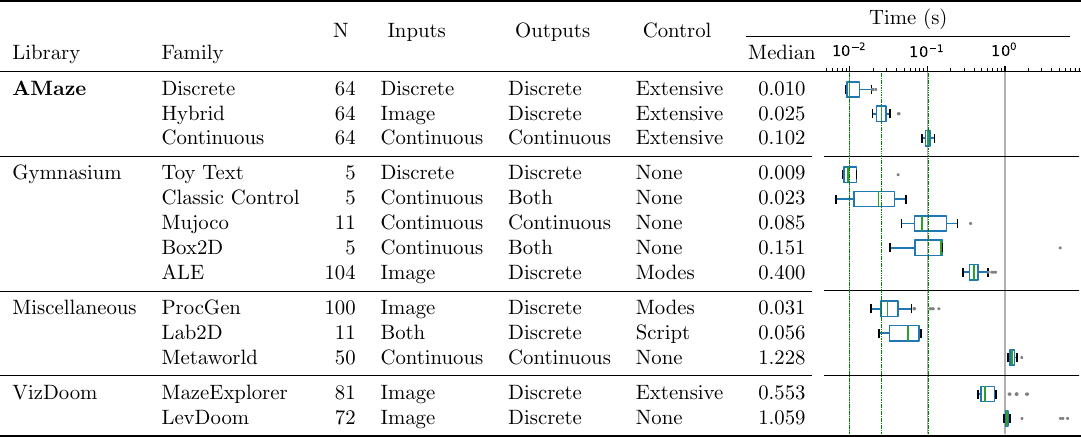}
\end{table}
}
\iftrailingfigures\else
\tabbenchmarks
\fi

To place this generator in perspective, we conducted an extensive comparison with a select number of commonly used benchmarks.
As our library is primarily targeted at Python environments, we restricted the set of considered environments to those that could be reliably installed and used on an experimenter's machine.
This led to the exclusion of the Unsupervised Reinforcement Learning Benchmark \citep{Laskin2021} and RetroGym \citep{Nichol2018} due to missing dependencies or tools.
On similar ground, Obstacle Tower \citep{Juliani2019} could not be evaluated here, despite the extensive control it gives the experimenter, as it required an independent Unity server to run, making it impractically slow.

As detailed in the following section, AMaze can provide environments for fully discrete, fully continuous, and hybrid agents.
\autoref{tab:benchmarks} illustrates how the former case allows for fast simulation at the cost of low observable complexity.
Based on the time taken to simulate 1000 timesteps, only the simplest of the gymnasium suite \citep{Sutton2018} is comparable to AMaze which, in addition, provides numerous unique and experimenter-controlled environments.
In the hybrid case, where agents perceive images but still only take discrete steps, the library is on par with Classic Control tasks \citep{Barto1983} such as Mountain Car or Cart Pole.
ProcGen \citep{Cobbe2020} addresses similar concerns as AMaze and is quite comparable in terms of speed, but has a stronger focus on randomness, with difficulty levels being the main way of controlling the resulting environments.
DeepMind Lab2D \citep{Beattie2020}, while noticably slower, is also extensively customizable, albeit through lua scripting, and allows for heterogeneous multi-agent experiments.

With respect to the fully discrete case, the most computationally expensive of the three regimes, AMaze performs at a level similar to that of Box2D or Mujoco \citep{Todorov2012}, which, in traditional implementations such as gymnasium \citep{Towers2024}, lack customization capabilities.
Purely vision-based benchmarks such as ALE \citep{Bellemare2013}, Meta-world \citep{Yu2019}, LevDoom \citep{Tomilin2022} or Maze Explorer \citep{Harries2019}, while offering a more challenging task than AMaze, also exhibit drastically higher costs with variable levels of experimenter control over the environments.

It follows that AMaze fills a very specific niche in the benchmarking landscape by providing a computationally inexpensive framework to design challenging environments.
Control over the various characteristics of said environments is left in the hands of the experimenter through a number of high- and low-level parameters that will be described in the following section.

\section{Generating mazes}
\label{sec:mazes}

Learning to navigate mazes represents a flexible, diverse, yet challenging testbed for training agents.
Here, we propose a \emph{generator} (AMaze) for this task with the following primary characteristics:

\def\figmodelmaze{
\begin{figure}[t]
 \centering
 \begin{minipage}{.5\columnwidth}
  \centering
  \input{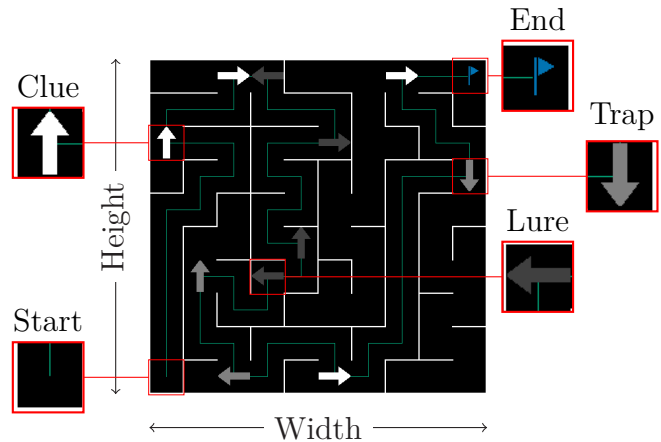}
 \end{minipage}
 \caption{Generic maze example. Agents start in one corner and must reach the opposite. Corridors can be empty or contain easily identifiable misleading signs (lures). Signs placed on intersections maybe trustworthy or not depending on whether they are a clue or a trap, respectively.}
 \label{fig:model:maze}
\end{figure}
}
\iftrailingfigures\else

\begin{figure}[t]
 \centering
 \begin{minipage}{.5\columnwidth}
  \centering
  \input{annexes/sample_maze.annotated}
 \end{minipage}
 \caption{Generic maze example. Agents start in one corner and must reach the opposite. Corridors can be empty or contain easily identifiable misleading signs (lures). Signs placed on intersections maybe trustworthy or not depending on whether they are a clue or a trap, respectively.}
 \label{fig:model:maze}
\end{figure}

\fi

\begin{itemize}
 \item [] \textbf{Loose embodiment} \par
 The agent has access only to local spatial information (its current cell) and limited temporal information (previous cell).
 Arbitrarily complex visual-like information is provided to the agent in either discrete (preprocessed) or continuous (image) form.
 \\

 \item [] \textbf{Computational lightweightness} \par
 No physics engine or off-screen renderings are required for such 2D mazes.
 Thus, challenging environments can be generated that are both observably complex \citep{Beattie2020} and relatively fast (as seen in \autoref{tab:benchmarks}).
 \\

 \item [] \textbf{Open-endedness} \par
 As illustrated in \autoref{fig:model:maze}, a given maze results from the interaction of numerous variables controlled by the experimenter, such as its dimensions or the frequency and type of visual cues.
 In practice, an experimenter can inject any level of complexity into the maze by selecting images of appropriate deceptiveness as cues.
\end{itemize}

These features make it possible to generate a wide range of mazes for a variety of purposes, from a fast prototyping RL platform to a testbed for embodied computer vision in indirectly encoded NeuroEvolution.
In the remainder of the section, we detail the major components of this generator namely the environment's parameters, the agents' capabilities, the reward function and, finally, unifying metrics for comparing widely different mazes.

\subsection{Maze generation}
A maze is defined, at its core, by its size (width, height) and the seed of a random number generator.
A depth-first search algorithm is then used to create the various paths and intersections with the arbitrary constraint that the final cell is always diagonally opposed to the starting point (itself a parameter).
Additionally, mazes can be made unicursive by blocking every intersection that does not lead directly to the goal.
Such mazes are called \emph{trivial}, as an optimal strategy simply requires going forward without hitting any wall.
In contrast, general-purpose mazes do have intersections, the correct direction being indicated by a sign, hereafter called a \emph{clue}.
This corresponds to the class of \emph{simple} mazes, since making the appropriate move in such cases is entirely context dependent.
Each intersection \emph{on the path to the goal} is labeled with such a clue.

However, to provide a sufficient level of difficulty, additional types of sign can also be added to a given maze with a user-defined probability.
\emph{Lures}, occurring with probability $p_l$, are easily identifiable erroneous signs that request an immediately unfavorable move (going backward or into a wall).
They can be placed on any nonintersectional cell along the path to the solution.
\emph{Traps}, replace an existing clue (with probability $p_t$) and instead point to a dead end.
These types of sign are much harder to detect as they do not violate local assumptions and can result in large, delayed negative rewards.
Mazes containing either of these misleading signs are named accordingly, while mazes containing \emph{both} are called \emph{complex}.

\subsection{Agents and state spaces}
To successfully navigate a maze, the learning agent must only rely on the visual contents of its current cell to choose its next action.
The framework accounts for three combinations of input/output types: fully discrete, fully continuous, and hybrid (continuous observations with discrete actions).
Observations in continuous space imply that cells are perceived directly as images albeit with a lower resolution than that presented to humans.
Thus, wall detection may not be initially trivial (even for unicursive mazes), and sign recognition comes into play with the possibility of using different symbols for different sign types.
In the discrete space, the agent is fed a sequence of eight floats, corresponding to preprocessed information in direct order ($W_e,W_n,W_w,W_s,S_e,S_n,S_w,S_s$), as illustrated in \autoref{fig:model:vision}.

\def\figmodelvision{
\begin{subfigure}
 
\tikzset{
  every node/.style={ultra thick},
  wnode/.style={solid},
  snode/.style={draw=blue}
 }
 \begin{minipage}[b]{.37\textwidth}
  \centering
  \begin{tikzpicture}[scale=.7]
   \foreach \r/\c/\L in {2/snode/S, 1/wnode/W} {
    \foreach \a/\l in {0/e, 90/n, 180/w, 270/s} {
     \draw [\c, fill=white, thick] (\a:\r) ++(-.5, -.5) rectangle ++(1, 1);
     \node at (\a:\r) {\footnotesize $\L_\l$};
    }
   }
  \end{tikzpicture}
  \caption{Visual inputs}
 \end{minipage}
 \begin{minipage}[b]{.57\textwidth}
  \centering%
  \def\i(#1,#2)#3#4{%
   \begin{scope}[shift={(10*#1, -10*#2)}]%
    \fill [gray] (-2.75, -2.75) rectangle ++(5.5, 5.5);
    \foreach \x [count=\c, evaluate=\c as \y using {{#4}[\c-1]}] in {0:1, 90:1, 180:1, 270:1, 0:2, 90:2, 180:2, 270:2} {
     \ifnum\c<5\def\d{wnode}\else\def\d{snode}\fi
     \fill [\y\relax, draw=black, \d, thick] (\x*1.1) ++(-.5, -.5) rectangle ++(1,1);
    }
    \node [anchor=north] at (0, -2.7) {\small #3};
   \end{scope}%
  }%
  \begin{tikzpicture}[scale=.21]
   \i(0,0){Start}{"white","black","white","white","black","black","black","black"}%
   \i(1,0){Clue}{"red","black","white","black","black","white","black","black"}%
   \i(2,0){End}{"white","white","red","black","black","black","black","black"}%

   \i(0,1){Trap}{"white","black","red","black","black","black","black","white!50!black"}%
   \i(1,1){Lure}{"black","white","white","black","black","black","white!25!black","black"}%

   \begin{scope}[shift={(20, -7)}, font=\footnotesize]
    \draw (-4.5, -.5) -- ++(10, 0);
    \draw (-4.5, -.5) -- ++(0, -6.5);
    \foreach \c/\l [count=\i] in {white/Wall, black/Empty, red/Origin, white/Sign} {
     \node [draw, fill=\c] (k\i) at (0, -1.5*\i) {}; \node (l\i) [right=0 of k\i] {\l};
    }
    \node [draw, snode, fill=black, left=0 of k2] {};
    \node [draw, snode, fill=white] at (k4) {};
    \node [draw, snode, fill=white!50!black, left=0 of k4] (k4_) {};
    \node [draw, snode, fill=white!25!black, left=0 of k4_] (k4__) {};
   \end{scope}
  \end{tikzpicture}
  \caption{Examples}
 \end{minipage}
 \setcounter{subfigure}{-1}

 \caption{Discrete observation space. a) $W_*$ denotes whether there is a wall in the corresponding direction, as well as the direction of the previous cell; $S_*$ is non-zero if a sign points towards the corresponding direction. b) Sample inputs from cells highlighted in \autoref{fig:model:maze}, as would be perceived by agents (without geometric relationship).}
 \label{fig:model:vision}
\end{subfigure}
}
\iftrailingfigures\else

\begin{subfigure}
 
\tikzset{
  every node/.style={ultra thick},
  wnode/.style={solid},
  snode/.style={draw=blue}
 }
 \begin{minipage}[b]{.37\textwidth}
  \centering
  \begin{tikzpicture}[scale=.7]
   \foreach \r/\c/\L in {2/snode/S, 1/wnode/W} {
    \foreach \a/\l in {0/e, 90/n, 180/w, 270/s} {
     \draw [\c, fill=white, thick] (\a:\r) ++(-.5, -.5) rectangle ++(1, 1);
     \node at (\a:\r) {\footnotesize $\L_\l$};
    }
   }
  \end{tikzpicture}
  \caption{Visual inputs}
 \end{minipage}
 \begin{minipage}[b]{.57\textwidth}
  \centering%
  \def\i(#1,#2)#3#4{%
   \begin{scope}[shift={(10*#1, -10*#2)}]%
    \fill [gray] (-2.75, -2.75) rectangle ++(5.5, 5.5);
    \foreach \x [count=\c, evaluate=\c as \y using {{#4}[\c-1]}] in {0:1, 90:1, 180:1, 270:1, 0:2, 90:2, 180:2, 270:2} {
     \ifnum\c<5\def\d{wnode}\else\def\d{snode}\fi
     \fill [\y\relax, draw=black, \d, thick] (\x*1.1) ++(-.5, -.5) rectangle ++(1,1);
    }
    \node [anchor=north] at (0, -2.7) {\small #3};
   \end{scope}%
  }%
  \begin{tikzpicture}[scale=.21]
   \i(0,0){Start}{"white","black","white","white","black","black","black","black"}%
   \i(1,0){Clue}{"red","black","white","black","black","white","black","black"}%
   \i(2,0){End}{"white","white","red","black","black","black","black","black"}%

   \i(0,1){Trap}{"white","black","red","black","black","black","black","white!50!black"}%
   \i(1,1){Lure}{"black","white","white","black","black","black","white!25!black","black"}%

   \begin{scope}[shift={(20, -7)}, font=\footnotesize]
    \draw (-4.5, -.5) -- ++(10, 0);
    \draw (-4.5, -.5) -- ++(0, -6.5);
    \foreach \c/\l [count=\i] in {white/Wall, black/Empty, red/Origin, white/Sign} {
     \node [draw, fill=\c] (k\i) at (0, -1.5*\i) {}; \node (l\i) [right=0 of k\i] {\l};
    }
    \node [draw, snode, fill=black, left=0 of k2] {};
    \node [draw, snode, fill=white] at (k4) {};
    \node [draw, snode, fill=white!50!black, left=0 of k4] (k4_) {};
    \node [draw, snode, fill=white!25!black, left=0 of k4_] (k4__) {};
   \end{scope}
  \end{tikzpicture}
  \caption{Examples}
 \end{minipage}
 \setcounter{subfigure}{-1}

 \caption{Discrete observation space. a) $W_*$ denotes whether there is a wall in the corresponding direction, as well as the direction of the previous cell; $S_*$ is non-zero if a sign points towards the corresponding direction. b) Sample inputs from cells highlighted in \autoref{fig:model:maze}, as would be perceived by agents (without geometric relationship).}
 \label{fig:model:vision}
\end{subfigure}

\fi

In this case, the observations take the form of a monodimensional array containing all eight fields, in direct order.
Signs can be differentiated through their associated decimal value, which is fully configurable by the experimenter.
In the subsequent experiment, we used a single sign of each type with values of 1.0, 0.5, and 0.25 for clues, traps, and lures, respectively.
The walls and the originating direction (limited temporal information) are assigned fixed values of 1.0 and 0.5, respectively.

With respect to actions, a discrete space implies that the agent moves directly from one cell to another by choosing one of the four cardinal directions.
In contrast, in a continuous action space, the agent controls only its acceleration.

\subsection{Reward function}

An optimal strategy, in the fully discrete case, is one where the agent makes no error: no wall collision, no backward steps, and, naturally, correct choices at all intersections.
Although identical in the hybrid case, as the increase in observation complexity does not change the fact that there exists only one optimal trajectory, this statement no longer holds for the fully continuous case, at least not in the trivial sense.
In fact, by controlling its acceleration, an agent can take shorter paths along corners or even take risks based on assumed corridor lengths.

However, in all cases, the same reward function is used to improve strategies as defined by:
\begin{equation}\begin{aligned}
 r(s, a, s') &= \rho_e, & \text{if } s' \text{ is the goal} \\
             &- \rho_w, & \text{if } a \text{ caused a collision} \\
             &- \rho_b, & \text{if } a \text{ caused a backward step} \\
             &- \rho_t, & \text{constant time penalty}
             \label{eq:reward}
\end{aligned}\end{equation}

Given $l$, the length of the optimal trajectory, we define two versions of the reward function: $r$ and its normalized version $\bar r$.
The first is used during the training process to provide large incentive towards reaching the goal, while the second's purpose is to compare performance on mazes with different sizes.
Furthermore, we refer to the cumulative (episodic) reward as $R$ and $\bar R$, respectively.
\autoref{tab:rewards} details the specific values used in this experiment.

\def\tabrewards{
\begin{table}[b]
 \centering
 \caption{Elementary rewards for both versions of the reward function (\autoref{eq:reward}): $r$ promotes reaching the goal with a large associated reward, while $\bar r$ always indicates an optimal strategy with a cumulative reward $\bar R = 1$. $l$ is the number of cells on the optimal path between start and finish.}
 \label{tab:rewards}
 \begin{tabular}{l*{4}{c}c}
           & $\rho_e$ & $\rho_w$ & $\rho_b$ & $\rho_t$  & Cumulative \\
 \midrule
       $r$ & $2l-1$   & $-0.1$   & $-0.2$   & $-1$      & $R = l$ \\
  $\bar r$ & $2$   & $-0.01$   & $-0.02$   & $-1/(l-1)$ & $\bar R = 1$ \\
 \end{tabular}
\end{table}
}
\iftrailingfigures\else
\tabrewards
\fi

\subsection{Evaluating maze complexity}

Due to the randomness of the generation process, two mazes with different seeds can have very different characteristics.
Thus, to provide a common ground from which mazes can be compared, we define two metrics based on Shannon's entropy \citep{Shannon1948}.
First the \textit{Surprisingness} $S(M)$ of a maze $M$:
\begin{equation}\label{eq:metric:suprisingness}
 S(M) = - \sum\limits_{i \in I_M} p(i) * log_2(p(i))
\end{equation}
where $p(i)$ is the observed frequency of input $i$ and $I_M$ is the set of inputs encountered when performing an optimal trajectory in $M$. Second, the \textit{Deceptiveness} $D(M)$ defined as:
\begin{align}\label{eq:metric:deceptiveness}
 \text{cells}(M) &= \{c[0:3], \forall c \in M\} \nonumber\\
 \text{traps}(M) &= \{c, \forall c \in M / \text{cost}(c) > 0\} \nonumber\\
 D(M) &= \sum\limits_{c \in \text{cells}(M)}
           \sum\limits_{\substack{s \in \text{traps}(M)\\s[0:3] = c}}
            - p(s|c) log_2(p(s|c))
\end{align}
where the cost of $c$ is above zero for cells containing traps and lures.

\def\figdistributions{
\begin{figure}
 \centering
 \input{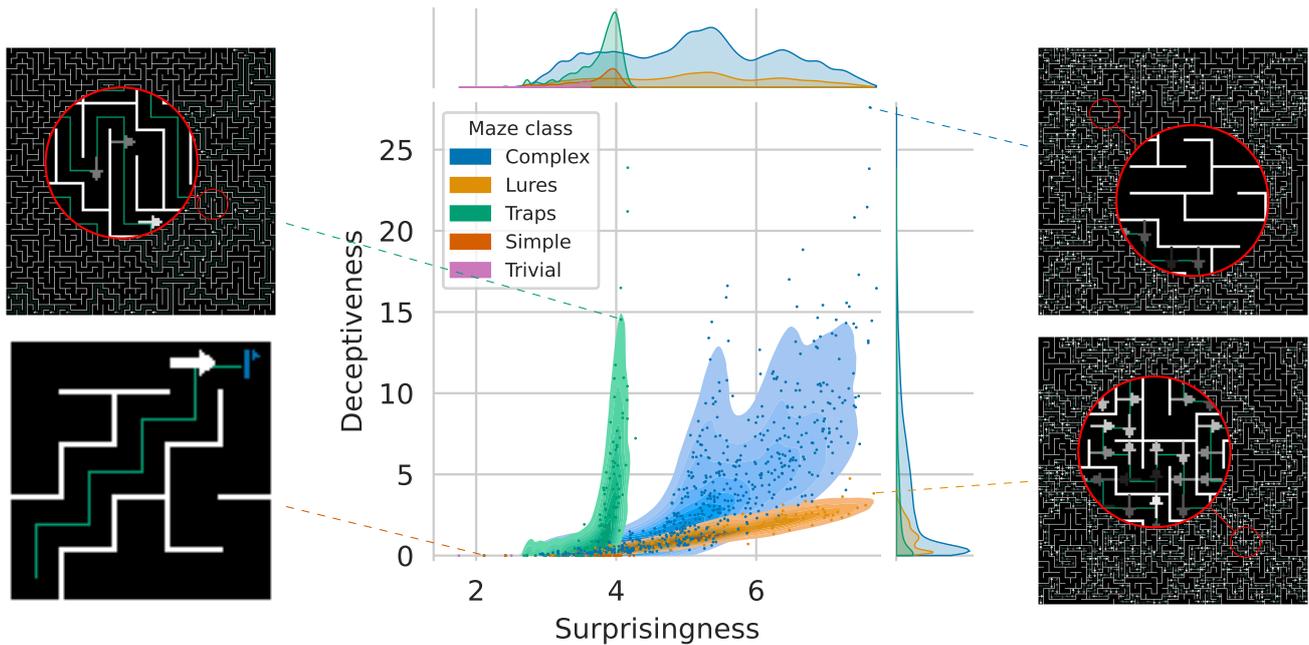}
 \caption{Distribution of Surprisingness versus Deceptiveness across 500'000 unique mazes from five different classes. The marginal densities for Surprisingness highlight the low number of different Trivial mazes ($[2,4]$ range), while classes of increasing difficulty allow for more variations. Examples of outlier mazes from the four main classes are depicted in the borders to illustrate the underlying Surprisingness (right column) or lack thereof (left column).}
 \label{fig:distributions}
\end{figure}
}
\iftrailingfigures\else

\begin{figure}
 \centering
 \input{annexes/complexity/distributions.annotated}
 \caption{Distribution of Surprisingness versus Deceptiveness across 500'000 unique mazes from five different classes. The marginal densities for Surprisingness highlight the low number of different Trivial mazes ($[2,4]$ range), while classes of increasing difficulty allow for more variations. Examples of outlier mazes from the four main classes are depicted in the borders to illustrate the underlying Surprisingness (right column) or lack thereof (left column).}
 \label{fig:distributions}
\end{figure}

\fi

As illustrated in \autoref{fig:distributions}, both metrics cover different regions of the maze space.
Surprisingness describes the likelihood of encountering numerous infrequent states while traversing the maze.
Conversely, the Deceptiveness focuses on the frequency with which deceptive states may be encountered, that is, it captures how ``dangerous'' the maze is.
One can see that, by taking advantage of both types of deceptive signs, Complex mazes exhibit the highest combined difficulty and frequency.
Furthermore, even with the limitations of discrete inputs, we can here see how it is theoretically possible to generate mazes of arbitrarily high Surprisingness and Deceptiveness.
Additional information and the data set on which these analyses are based can be found in the associated Zenodo record \citep{GodinDubois2024za}.

\section{Training protocol on AMaze}
\label{sec:training}

To teach agents generalizable navigation skills, we define a \emph{training} maze, presented in \autoref{fig:protocol:direct:train}.
Although, for simplicity, we only depict one variation of this maze, in practice, the agent is trained on all four rotations (\autoref{fig:protocol:direct:rotations}).
Thus, the agent will not overfit to a particular upper-diagonal type of behavior, but instead will have to develop a context-dependent strategy.
Furthermore, intermediate evaluations of the agent's performance are performed in a similar maze (in terms of complexity) with a different seed, as shown in \autoref{fig:protocol:direct:eval}.

\def\figprotocoldirect{
\begin{subfigure}
 \centering
\hfill
\foreach \k/\c/\id/\m in {Training/train/1/07, Evaluation/eval/0/18} {
 \begin{minipage}[b]{.3\columnwidth}
  \includegraphics[width=\columnwidth]{annexes/data/interpolation/\id_9_M20000\m_20x20_C1_l.25_L.25_t.5_T.5}%
  \caption{\k\label{fig:protocol:direct:\c}}
 \end{minipage}%
\hfill
}%
\begin{minipage}[b]{.3\columnwidth}
 \begin{tikzpicture}
  \foreach \a [count=\i] in {0,90,180,270} {
   \node at (0, 0) [anchor=45+\a, inner sep=0pt]
    {\includegraphics[width=.5\textwidth, angle=\a, origin=c]{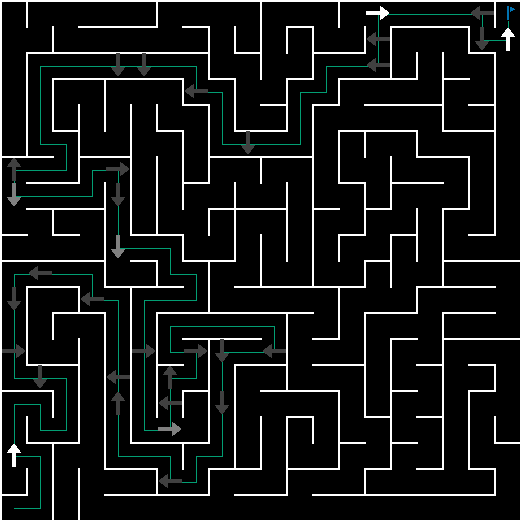}};
  }
 \end{tikzpicture}
 \caption{All rotations\label{fig:protocol:direct:rotations}}
\end{minipage}%
\setcounter{subfigure}{-1}
\hfill

 \caption{Mazes used in direct training. a) maze used to collect experiences and learn from. b) maze used to periodically evaluate performance. Note that, in practice, the agent experiences mazes as in (c), i.e., with all rotations for both training and evaluations.}
 \label{fig:protocol:direct}
\end{subfigure}
}
\iftrailingfigures\else

\begin{subfigure}
 \centering
\hfill
\foreach \k/\c/\id/\m in {Training/train/1/07, Evaluation/eval/0/18} {
 \begin{minipage}[b]{.3\columnwidth}
  \includegraphics[width=\columnwidth]{annexes/data/interpolation/\id_9_M20000\m_20x20_C1_l.25_L.25_t.5_T.5}%
  \caption{\k\label{fig:protocol:direct:\c}}
 \end{minipage}%
\hfill
}%
\begin{minipage}[b]{.3\columnwidth}
 \begin{tikzpicture}
  \foreach \a [count=\i] in {0,90,180,270} {
   \node at (0, 0) [anchor=45+\a, inner sep=0pt]
    {\includegraphics[width=.5\textwidth, angle=\a, origin=c]{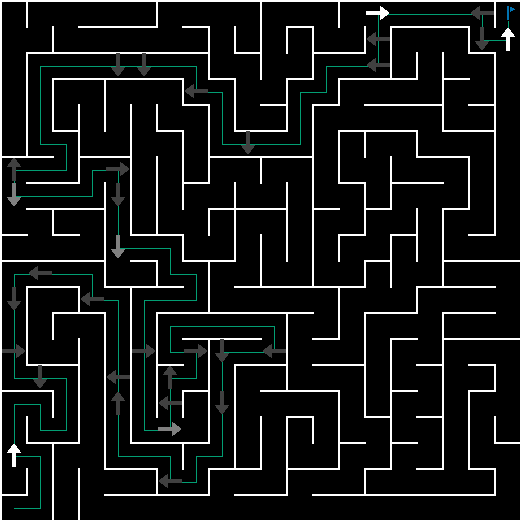}};
  }
 \end{tikzpicture}
 \caption{All rotations\label{fig:protocol:direct:rotations}}
\end{minipage}%
\setcounter{subfigure}{-1}
\hfill

 \caption{Mazes used in direct training. a) maze used to collect experiences and learn from. b) maze used to periodically evaluate performance. Note that, in practice, the agent experiences mazes as in (c), i.e., with all rotations for both training and evaluations.}
 \label{fig:protocol:direct}
\end{subfigure}

\fi

To showcase this benchmark's integration with current Reinforcement Learning frameworks, we used Stable Baselines 3 \citep{Raffin2021} and more specifically their off-the-shelf Advantage Actor-Critic (A2C) and Proximal Policy Optimization (PPO) algorithms with all hyperparameters kept to their default values\citep{Mnih2016, Schulman2017}.
The total training budget is of 3000000 timesteps, divided over the four rotational variations of the training maze with possible early stopping if the optimal trajectory is observed on all evaluation mazes.

\subsection{Scaffolding}

In addition to this \emph{direct} training in a hard maze, we also followed two incremental protocols: an \emph{interpolation} training, which ``smoothly'' transitions from simple to more complex mazes, and the \emph{EDHuCAT} training, which leverages human creativity and reactivity \citep{Eiben2015}.
In the former case, agents start from trivial environments and gradually move onto harder challenges.
However, the final mazes on which agents are trained and evaluated are identical to those of the direct case.
Succinctly, every atomic parameter is interpolated between the initial and final mazes' values according to specific per-field rules, e.g. for the apparition of intersections or traps.
A total of ten training stages are performed in this protocol, that is 300'000 timesteps per stage.
In case of early convergence, the remainder of the budget is transferred equally to future stages.

\subsection{Interactive training}

In the interactive setup, we use the Environment-Driven Human-Controlled Automated Training (EDHuCAT) algorithm, loosely inspired by the EDEnS\footnote{Environment-Driven Evolutionary Selection \citep{GodinDubois2020a}, used for automated open-ended evolution} algorithm.
As summarized in \autoref{alg:edhucat}, EDHuCAT operates under the joint principles of concurrency (multiple agents evaluated in parallel) and diversity (multiple mazes are generated by the user/experimenter).
The advantage of this method over the simple interpolation between initial and final mazes is that it can take advantage of unforeseen developments that occur in the middle of the training.
For instance, if the human agent detects that the learning agent has too much difficulty with some newly presented features, they can decide to decrease the difficulty, select from a wider diversity of mazes, or even increase the difficulty.
At the same time, the human component makes it harder for the training algorithm itself (A2C or PPO) due to the potential introduction of so-called moving targets (see \autoref{sec:results:human}).
That is, a Human may not follow a strict policy for choosing mazes or agents, whether between replicates or even during a given run.
The total budget is the same as for the other protocols; however, as three concurrent evaluations are performed for each stage, an agent in a given stage is only trained for a maximum of 100000 time steps.

\begin{algorithm2e}[t]
 \def\HiLi{}
 \def\HiLi{%
  \leavevmode\rlap{%
   \hbox to \hsize{%
    \color{yellow!50}%
    \leaders%
    \hrule height .8%
    \baselineskip depth .5ex%
    \hfill%
   }%
  }%
 }%

 \SetKwInOut{Input}{input}\SetKwInOut{Output}{output}
\Input{K, Number of concurrent environments\\
        S, number of intermediate steps\\
        m, an initial maze\\
        a human}
\Output{EDHuCAT\emph{ed} agent}
a $\leftarrow$ Agent()\;
train(a, m)\;
agents $\leftarrow$ [a]\;
\For{$i\leftarrow 1$ \KwTo $S-1$}{
 \HiLi a $\leftarrow$ select(agents)\;
 \HiLi mazes $\leftarrow$ generate(K)\;
 agents $\leftarrow$ copy(a, K)\;
 \For{$k\leftarrow 1$ \KwTo $K$}{
  train(agents[k], mazes[k])\;
 }
}
\KwRet select(agents)\;

 \caption{\edhucat{} algorithm. A human agent is used to perform the \emph{select} and \emph{generate} operations. In this work $K=3$ and $S=10$.}
 \label{alg:edhucat}
\end{algorithm2e}

\section{Evaluation of generalized performance}
\label{sec:eval}

\def\figevalmazes{
\begin{figure}
 \centering
\begin{tikzpicture}
 \begin{localtoimage}{reeval/mazes}{width=.9\textwidth}
  \draw [->] (-1.1, 1.1) -- ++(0, -2.1) node [rotate=90, fill=white, pos=.5] {More complex};
  \draw [->] (-1.1, 1.1) -- ++(2.1, 0) node [fill=white, pos=.5] {More features};
  \draw [->] (0, 1.2) -- ++(1, 0) node [fill=white, pos=.5] {More traps};
  \foreach \l [count=\i] in {Trivial, Simple, Lures, 1 Trap, 3 Traps, 16 Traps} {
   \pgfmathsetmacro{\x}{10/6*(\i-1)/5-5/6}
   \node at (\x,-1) [anchor=north] {\small\l};
  }
 \end{localtoimage}
\end{tikzpicture}
 \caption{Mazes used for generalization evaluation. The first three columns correspond to different maze classes, while the last three all include traps but with different frequencies (1, 3, 16). Each row corresponds to the minimal, median, and maximal complexity of mazes obtained from a random sample of size 10000.}
 \label{fig:eval:mazes}
\end{figure}
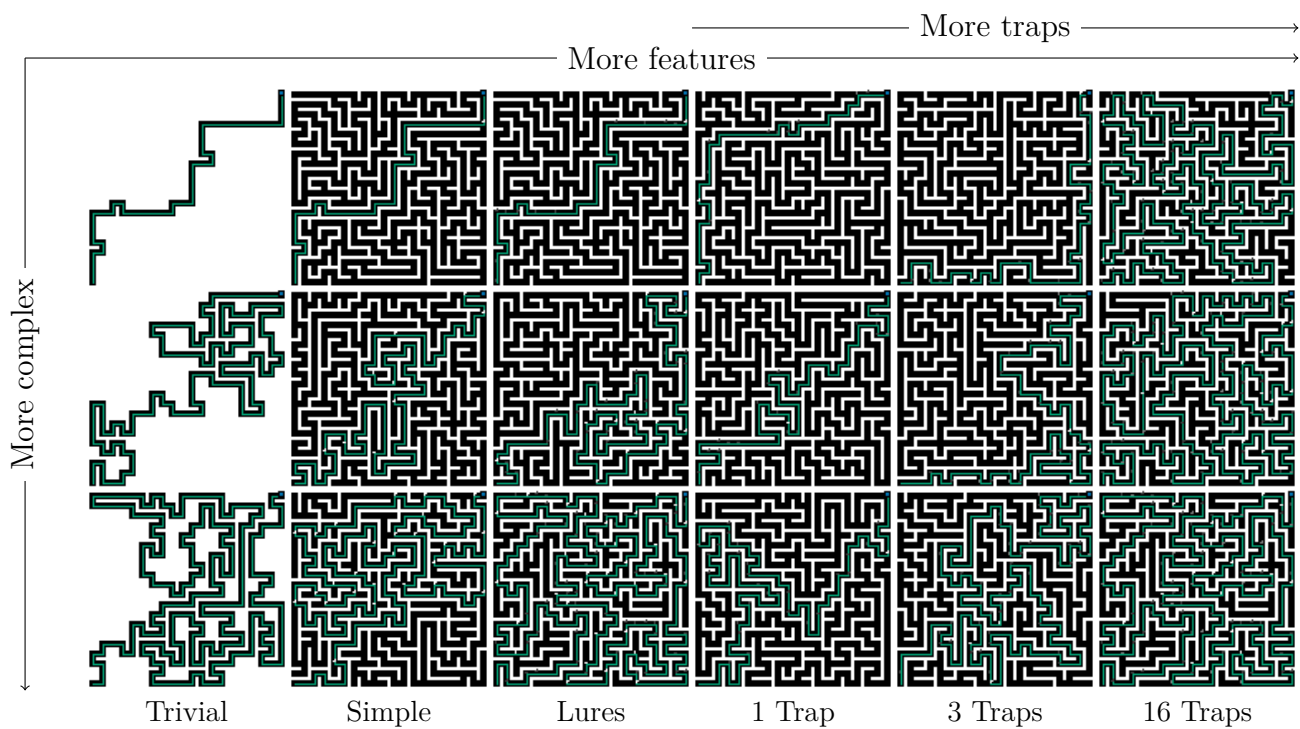
}
\iftrailingfigures\else

\begin{figure}
 \centering
\begin{tikzpicture}
 \begin{localtoimage}{reeval/mazes}{width=.9\textwidth}
  \draw [->] (-1.1, 1.1) -- ++(0, -2.1) node [rotate=90, fill=white, pos=.5] {More complex};
  \draw [->] (-1.1, 1.1) -- ++(2.1, 0) node [fill=white, pos=.5] {More features};
  \draw [->] (0, 1.2) -- ++(1, 0) node [fill=white, pos=.5] {More traps};
  \foreach \l [count=\i] in {Trivial, Simple, Lures, 1 Trap, 3 Traps, 16 Traps} {
   \pgfmathsetmacro{\x}{10/6*(\i-1)/5-5/6}
   \node at (\x,-1) [anchor=north] {\small\l};
  }
 \end{localtoimage}
\end{tikzpicture}
 \caption{Mazes used for generalization evaluation. The first three columns correspond to different maze classes, while the last three all include traps but with different frequencies (1, 3, 16). Each row corresponds to the minimal, median, and maximal complexity of mazes obtained from a random sample of size 10000.}
 \label{fig:eval:mazes}
\end{figure}

\fi

Following the training protocols defined in the previous section, we evaluated the final agents on two complementary tasks to determine whether they had acquired generalized behavior in the target maze class.
The first is straightforward: can the agent solve any maze of a given complexity or lower?
To answer this, we generated 18 mazes, as shown in \autoref{fig:eval:mazes}, based on varying amounts of features (clues, lures, traps) and Surprisingness.
The agents are then evaluated with respect to two goals: their success (do they reach the goal) and their reward (cumulative normalized reward $\bar R$, as in \autoref{eq:reward}).
Although this allows for comparison between agents based on performance under ``normal conditions'', this method suffers from cumulative failure: an error at a given time point may preclude any further success.
In fact, agents who take a wrong turn somewhere have little information on how to get back on track.
Thus, suboptimal strategies may end up indistinguishable from trivially bad ones.

To counteract this trend, we also perform a complementary evaluation in a more abstract context.
Because the input is discrete and thus enumerable, we can generate the complete set of possible input arrays.
As we know which is the correct decision, we can assess which inputs are correctly processed by the agents among the four classes: \emph{empty} corridor, corridor with \emph{lure}, intersection with \emph{clue}, and intersection with \emph{trap}.
Although less ``natural'', this method ensures complete coverage of all the possible situations that an agent may encounter on an infinite number of mazes.
Conversely, it also implies that we may be testing an agent on input configurations that it has never seen during training.

\subsection{Generalized maze-navigation}

\def\figresultsmazes{
\begin{subfigure}
 \begin{minipage}{.49\textwidth}
 \includegraphics[width=\textwidth, page=1]{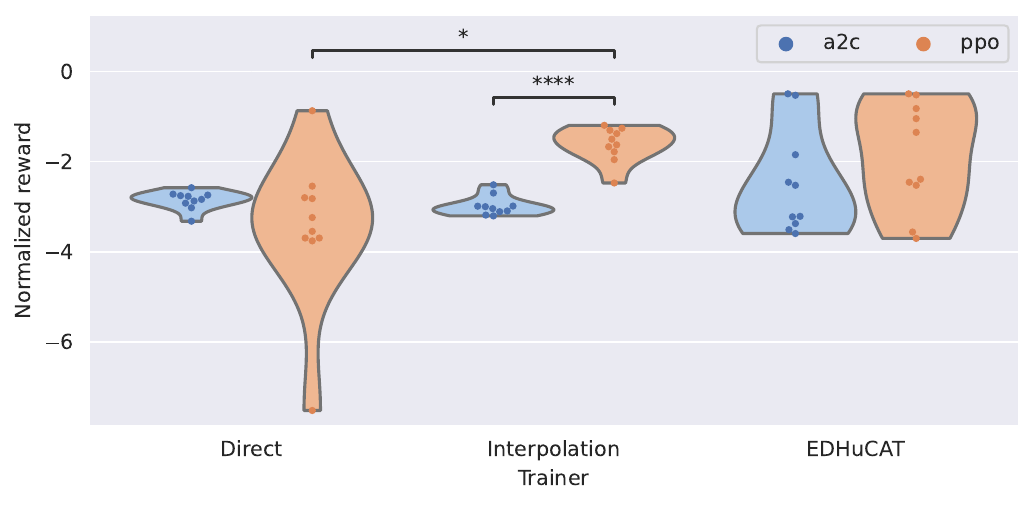}
 \caption{Average normalized reward}
 \label{fig:results:mazes:rewards}
\end{minipage}
\begin{minipage}{.49\textwidth}
 \includegraphics[width=\textwidth, page=2]{results}
 \caption{Average success rate}
 \label{fig:results:mazes:success}
\end{minipage}
\setcounter{subfigure}{-1}

 \caption{Normalized rewards and maze completion rates across trainers and algorithms. (a) \edhucat{} is better than direct training, but more dispersed than interpolation. (b) PPO is dramatically better for interpolation, while its advantage with \edhucat{} is unclear. Statistical differences were obtained with an independent t test and a Benjamini-Hochberg correction. *: p-value < 0.05, ****: p-value < .0001.}
 \label{fig:results:mazes}
\end{subfigure}
}
\iftrailingfigures\else

\begin{subfigure}
 \begin{minipage}{.49\textwidth}
 \includegraphics[width=\textwidth, page=1]{results}
 \caption{Average normalized reward}
 \label{fig:results:mazes:rewards}
\end{minipage}
\begin{minipage}{.49\textwidth}
 \includegraphics[width=\textwidth, page=2]{results}
 \caption{Average success rate}
 \label{fig:results:mazes:success}
\end{minipage}
\setcounter{subfigure}{-1}

 \caption{Normalized rewards and maze completion rates across trainers and algorithms. (a) \edhucat{} is better than direct training, but more dispersed than interpolation. (b) PPO is dramatically better for interpolation, while its advantage with \edhucat{} is unclear. Statistical differences were obtained with an independent t test and a Benjamini-Hochberg correction. *: p-value < 0.05, ****: p-value < .0001.}
 \label{fig:results:mazes}
\end{subfigure}

\fi

As summarized in \autoref{fig:results:mazes}, the average cumulative rewards ($\bar R$) and the success rates (fraction of mazes whose target was reached) are uniquely distributed according to the training regimen and algorithm.
For rewards, both direct and interpolation training have similar trends when using the A2C algorithm (in $[-2, -4]$), while \edhucat{} stands out with a more dispersed distribution.
When considering the PPO algorithm, there is a clear negative impact of direct training versus both alternatives.
Although \edhucat{} still presents a higher variance than interpolation training, both generated agents who obtained better rewards.
Furthermore, in the latter case, PPO significantly outperforms A2C with a p-value < 0.0001 for an independent t-test with Benjamini-Hochberg correction \citep{Benjamini1995}.

This difference is more clearly visible with the maze completion rate (\autoref{fig:results:mazes:success}), especially for agents generated by interpolation training: the best A2C agent is comparable to the worst PPO agent.
Again, this is strongly confirmed statistically using the same methodology and with a similar p-value.
Additionally, it would seem, from this distribution, that A2C is a slightly better choice in static environments (direct/A2C is marginally better than interpolation/A2C) and conversely for dynamical environments (direct/PPO generally lower than interpolation/PPO).
As previously, the human interventions promoted by \edhucat{} do not appear to be beneficial to the PPO algorithm.

\subsection{Generalized input-processing}

\def\figresultsinputs{
\setcounter{figure}{7}
\begin{subfigure}
 \begin{minipage}[b]{.49\textwidth}
 \includegraphics[width=\textwidth, page=3]{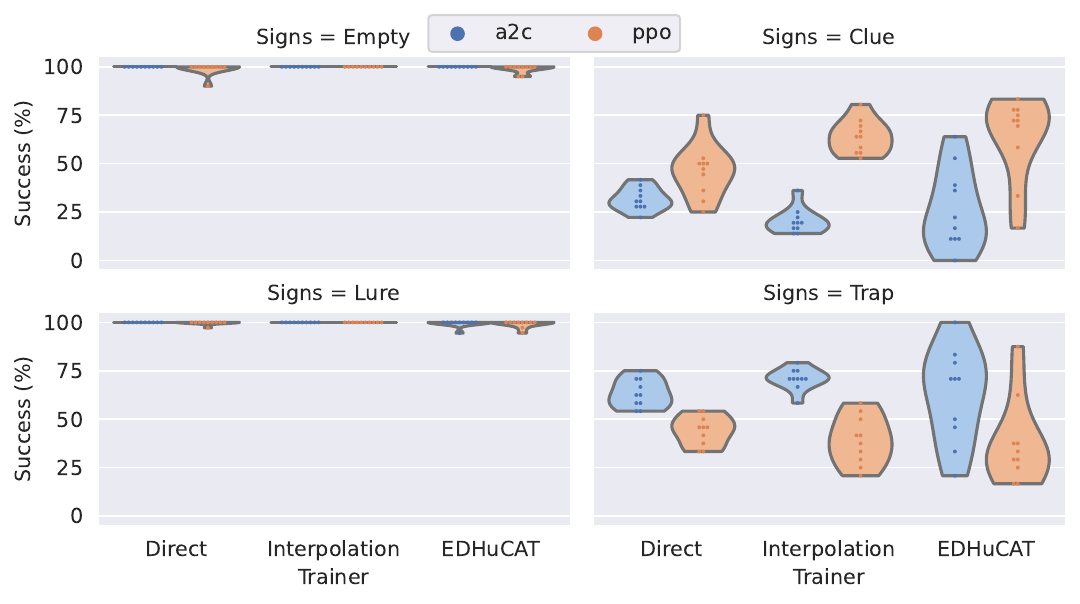}
 \caption{Per sign type}
 \label{fig:results:inputs:details}
\end{minipage}
\begin{minipage}[b]{.49\textwidth}
 \includegraphics[width=\textwidth, page=4]{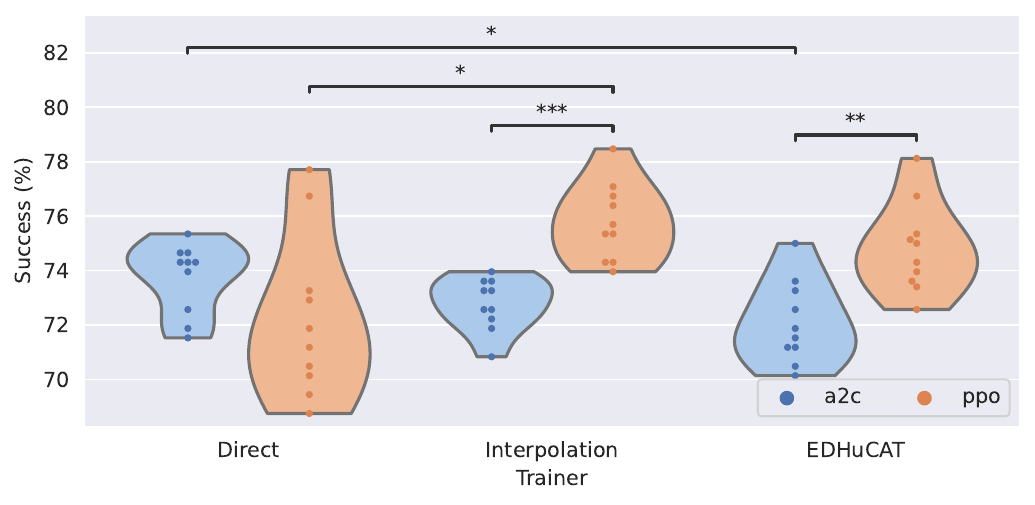}
 \caption{Average performance}
 \label{fig:results:inputs:average}
\end{minipage}
\setcounter{subfigure}{-1}

 \caption{Correct input processing rate across trainers and algorithms. (a) Corridors (with and without lures) are trivial, while A2C detects traps more efficiently than PPO and the other way around. (b) PPO outperforms A2C except for direct training, while nonstationary training seems overall beneficial. Statistical differences were also obtained with an independent t-test and a Benjamini-Hochberg correction: $0.05 \le$ * $< 10^{-2} \le$ ** $< 10^{-3} \le$ *** $< 10^{-4} \le$ ****.}
 \label{fig:results:inputs}
\end{subfigure}
}
\iftrailingfigures\else

\setcounter{figure}{7}
\begin{subfigure}
 \begin{minipage}[b]{.49\textwidth}
 \includegraphics[width=\textwidth, page=3]{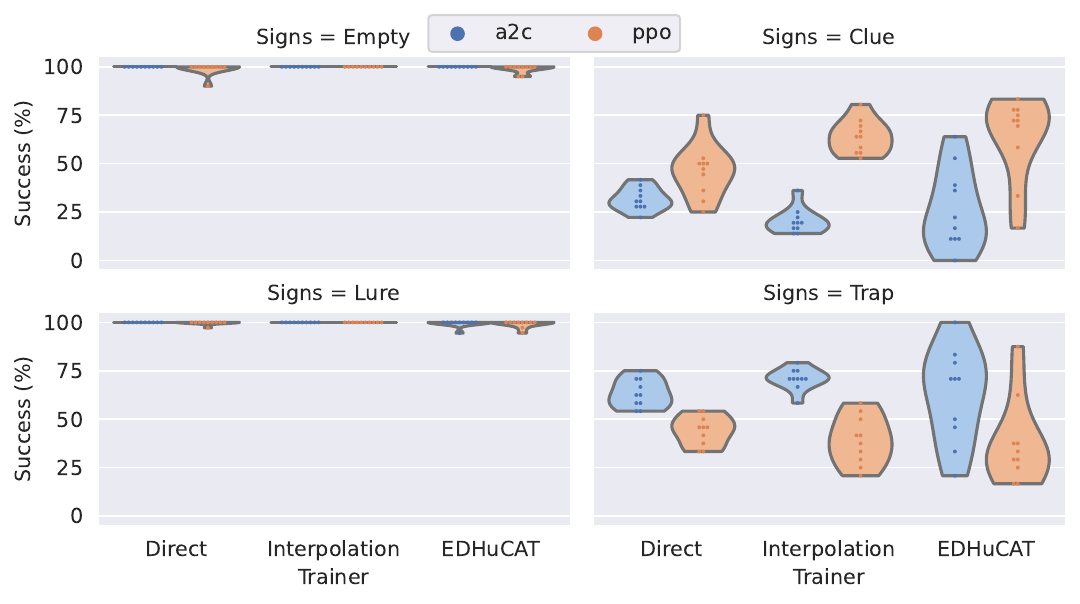}
 \caption{Per sign type}
 \label{fig:results:inputs:details}
\end{minipage}
\begin{minipage}[b]{.49\textwidth}
 \includegraphics[width=\textwidth, page=4]{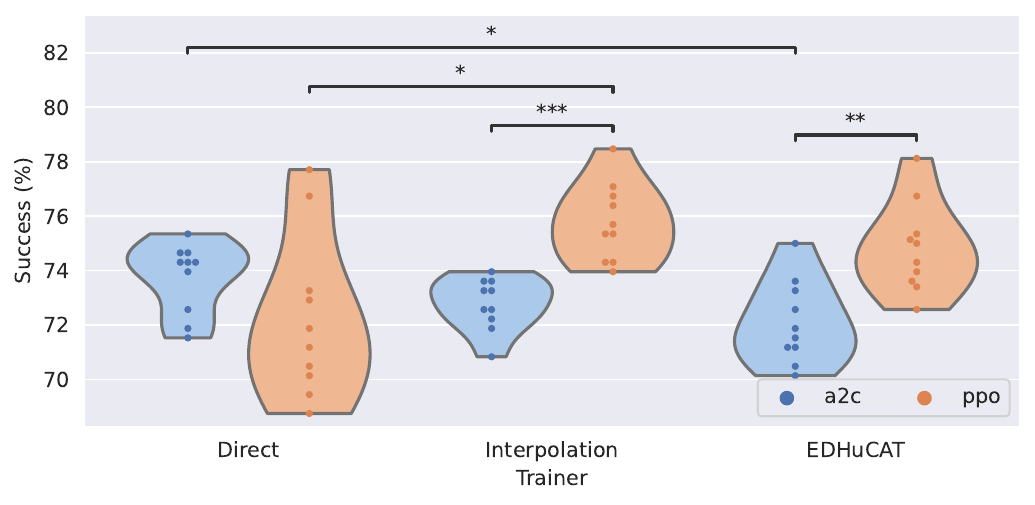}
 \caption{Average performance}
 \label{fig:results:inputs:average}
\end{minipage}
\setcounter{subfigure}{-1}

 \caption{Correct input processing rate across trainers and algorithms. (a) Corridors (with and without lures) are trivial, while A2C detects traps more efficiently than PPO and the other way around. (b) PPO outperforms A2C except for direct training, while nonstationary training seems overall beneficial. Statistical differences were also obtained with an independent t-test and a Benjamini-Hochberg correction: $0.05 \le$ * $< 10^{-2} \le$ ** $< 10^{-3} \le$ *** $< 10^{-4} \le$ ****.}
 \label{fig:results:inputs}
\end{subfigure}

\fi

We can make similar observations for the direct input processing test, as illustrated in \autoref{fig:results:inputs}.
Selecting the correct action is almost perfectly done by all agents, across all treatments, for the simplest cases (empty corridors and corridors with lures).
Surprisingly, the reaction to the presence of a nontrivial sign is handled differently depending on the algorithm.
Although PPO seems to be more efficient in detecting clues, A2C shows a better response to traps (\autoref{fig:results:inputs:details}).
Nonetheless, we can see that, on average, PPO shows clear benefits over A2C (\autoref{fig:results:inputs:average}).
With this test, we can confirm the advantage of using the former over the latter when facing dynamic environments.
The statistical significance is lower than $10^{-3}$ and $10^{-2}$ for the interpolation training and \edhucat{}, respectively.
In contrast, there is a marginally significant negative trend between A2C use and environmental variability.

\subsection{Aggregated performance}

\def\tabperformance{
\begin{table}
 \centering
 \caption{Aggregated maximum and median performance by trainer and algorithm. The best values for a row are indicated in \textbf{bold} and second-best in \textit{italic}. \edhucat{} produced the most general maze navigation agent with respect to normalized rewards and maze completion rates (top rows). Interpolation and \edhucat{} show complementary capabilities to produce better maze navigation \emph{in general} (bottom rows).}
 \label{tab:performance}
 \let\b\bfseries
\let\i\itshape
\begin{tabular}{llcccccc}
\toprule
 \multirow{2}{*}{}
 & Trainer           & \multicolumn{2}{c}{Direct}
                                         & \multicolumn{2}{c}{Interpolation}
                                                           & \multicolumn{2}{c}{EDHuCAT} \\
 & Algorithm         & A2C    & PPO      & A2C    & PPO      & A2C    & PPO     \\
\midrule
 \multirow{3}{*}{Maximum}
 & Normalized reward & -2.58  & \i-0.873 & -2.51  & -1.2     & \b-0.498 & \b-0.498 \\
 & Success rate      &  0.347 &    0.778 &  0.403 &  0.708   &  \i0.792 &  \b0.806 \\
 & Input recognition &  0.753 &    0.777 &  0.74  & \b0.785  &    0.75  &  \i0.781 \\
\midrule
 \multirow{3}{*}{Median}
 & Normalized reward & -2.8   &   -3.39  & -3.02  & \b-1.57  &   -2.87  & \i-1.87  \\
 & Success rate      &  0.34  &    0.326 &  0.257 &  \i0.618 &    0.514 &  \b0.667 \\
 & Input recognition &  0.743 &    0.715 &  0.729 &  \b0.755 &    0.717 &  \i0.747 \\
\bottomrule
\end{tabular}

\end{table}
}
\iftrailingfigures\else

\begin{table}
 \centering
 \caption{Aggregated maximum and median performance by trainer and algorithm. The best values for a row are indicated in \textbf{bold} and second-best in \textit{italic}. \edhucat{} produced the most general maze navigation agent with respect to normalized rewards and maze completion rates (top rows). Interpolation and \edhucat{} show complementary capabilities to produce better maze navigation \emph{in general} (bottom rows).}
 \label{tab:performance}
 \let\b\bfseries
\let\i\itshape
\begin{tabular}{llcccccc}
\toprule
 \multirow{2}{*}{}
 & Trainer           & \multicolumn{2}{c}{Direct}
                                         & \multicolumn{2}{c}{Interpolation}
                                                           & \multicolumn{2}{c}{EDHuCAT} \\
 & Algorithm         & A2C    & PPO      & A2C    & PPO      & A2C    & PPO     \\
\midrule
 \multirow{3}{*}{Maximum}
 & Normalized reward & -2.58  & \i-0.873 & -2.51  & -1.2     & \b-0.498 & \b-0.498 \\
 & Success rate      &  0.347 &    0.778 &  0.403 &  0.708   &  \i0.792 &  \b0.806 \\
 & Input recognition &  0.753 &    0.777 &  0.74  & \b0.785  &    0.75  &  \i0.781 \\
\midrule
 \multirow{3}{*}{Median}
 & Normalized reward & -2.8   &   -3.39  & -3.02  & \b-1.57  &   -2.87  & \i-1.87  \\
 & Success rate      &  0.34  &    0.326 &  0.257 &  \i0.618 &    0.514 &  \b0.667 \\
 & Input recognition &  0.743 &    0.715 &  0.729 &  \b0.755 &    0.717 &  \i0.747 \\
\bottomrule
\end{tabular}

\end{table}

\fi

To better compare the general performance of all training regimens and algorithms, we provide the maximum and median performance of the six combinations for the three metrics in \autoref{tab:performance}.
\edhucat{} succeeded in generating the most general maze navigator of all treatments with an average normalized reward of -0.498, compared to -0.873 and -1.2 of direct and interpolation trainings, respectively.
Surprisingly, such rewards were obtained with both algorithms, while alternatives fared much worse when using A2C.
Furthermore, it reaches a maze completion rate of 80.6\% with PPO and 79.2\% with A2C, again taking the lead on direct training (77. 8\%).
Interpolation showed more promise with the input recognition metric, although the low overall variations of this metric preclude additional inferences.

Complementarily, in the context of easily generating general maze-navigating agents, the median performance is useful to highlight which combination of training regime and algorithm was better across replicates.
Although slightly less favorable for \edhucat, which is in the top position once and second position twice, the results still speak volumes in favor of nonstationary environments.
However, this time around, PPO is clearly identifiable as the algorithm that performs the best, since \edhucat{} also shows a marked bias in its favor.

\subsection{Human impact}
\label{sec:results:human}

The previous metrics showed how agents resulting from the \edhucat{} algorithm can have a wide range of performance.
To provide a tentative investigation of the reasons for this variability, we classified the decisions made by the human agent into three categories: \emph{Careful}, challenges are slowly integrated once previous ones are solved; \emph{Risky}, the agent is exposed to unfair conditions to promote resilience; \emph{Moderate}, new challenges can be presented even if the agent has not solved the previous ones.
The results (in the associated record) show that the \emph{Careful} strategy provides better performance.
Agents resulting from both the PPO algorithm and this strategy often end at the top, while agents training with A2C followed an inverse trend.

\section{Conclusion and discussion}

In this work, we presented a benchmark generator that is geared toward the easy generation of feature-specific mazes and the intuitive understanding of the resulting agents' strategies.
The visual cues (either pre-processed or raw) these agents must learn to use to successfully navigate mazes are designed in a CPU-friendly manner so as to drastically limit computational time.
By grounding an embodied visual task in what is essentially a succession of lookup-table queries, we allow complex cognitive processes to take place while avoiding the cost of a full robotics simulator.
As the agents have only access to local information, this generator is applicable across a broad range of research domains, e.g. from sequential decision making to embodied AI.
To help future researchers in manipulating and comparing mazes with widely different characteristics, we introduced two partially orthogonal metrics that accurately capture two key features of such mazes: their Surprisingness and Deceptiveness.

Furthermore, to demonstrate the potential of this generator, we compared the training capabilities of the Advantage Actor-Critic (A2C) and Proximal Policy Optimization (PPO) algorithms in three different training regimens with varying levels of environmental diversity.
Direct training was a brute-force approach with only a target maze, while the Interpolation case relied on a scaffolding approach presenting increasing challenges.
Finally, an interactive methodology (\edhucat) was introduced to leverage human expertise as often as possible.

We evaluated the performance of both the maze navigation capabilities of trained agents and their ability to correctly process the entire observation space.
Across all these metrics, it was shown that PPO significantly outperforms A2C in dynamic environments, demonstrating the relevance of the former in producing generalized agents.
Furthermore, we found that \edhucat{} together with PPO was clearly one step above the alternatives when aiming for \emph{a} general maze-navigating agent.
At the same time, if one strives for more than a singular champion but, instead, for reproducible performance, then results point to both the Interpolation and interactive training setups as valid contenders when used in conjunction with PPO.

While demonstrating the potential of AMaze as a benchmark generator for AI agents, this work also raised a number of questions.
First, we aim to confirm whether the observed higher performance of PPO is explained by its use of a trust region, which reduces learning speed and, in turn, overfitting.
Furthermore, as we limited the study to two RL algorithms and a single neural architecture, many questions remain open with respect to the best choice of hyperparameters or even the applicability of other techniques, such as Evolutionary Algorithms.
Second, we only briefly mentioned the impact of the human in the interactive case, and while preliminary data \citep{GodinDubois2024za} show tentative relationships between the human strategy, the training algorithm, and performance, dedicated studies are required to provide definitive answers.
The strategy could be studied, as well as additional factors: Do youngsters train better than their elders? Does having a background in AI help? Or can laymen outperform experts?

\section{Acknowledgments}

This research was funded by the Hybrid Intelligence Center, a 10-year programme funded by the Dutch Ministry of Education, Culture and Science through the Netherlands Organisation for Scientific Research, \url{https://hybrid-intelligence-centre.nl}, grant number 024.004.022.

\bibliographystyle{Frontiers-Harvard}
\bibliography{main}

\end{document}